\documentclass[runningheads]{llncs}
\usepackage{graphicx}
\usepackage{float}
\usepackage{caption, copyrightbox}
\usepackage{subcaption}
\usepackage{algorithmic}
\usepackage{algorithm}
\usepackage{amsmath}

\newcommand{\source}[1]{\caption*{Source: {#1}} }
\begin{document}
\title{Effect of Superpixel Aggregation on Explanations in LIME -- A Case Study with Biological Data
}

\titlerunning{Effect of Superpixel Aggregation on Explanations in LIME}
%
\author{Ludwig Schallner$^1$
\and
Johannes Rabold$^2$
‌\and
Oliver Scholz$^1$
\and
Ute Schmid$^2$
}
\authorrunning{L. Schallner, et al.}
%
\institute{Fraunhofer IIS/EZRT, Germany \and
Cognitive Systems, University of Bamberg, Germany
}
\maketitle              
\begin{abstract}
	End-to-end learning with deep neural networks, such as convolutional
	neural networks (CNNs), has been demonstrated to be very successful for different tasks of image classification.	To make decisions of black-box approaches transparent, different solutions have been proposed. LIME is an approach to explainable AI relying on segmenting images into superpixels based on the Quick-Shift algorithm. In this paper, we present an explorative study of how different superpixel methods, namely Felzenszwalb, SLIC and Compact-Watershed,  impact the generated visual explanations. We compare the resulting relevance areas with the image parts marked by a human reference. Results show that image parts selected as relevant strongly vary depending on the applied method. Quick-Shift resulted in the least and Compact-Watershed in the highest correspondence with the reference relevance areas.
	

\end{abstract}
\section{Introduction}
Especially in visual domains, deep Convolutional Neural Networks (CNNs) have shown their superior capabilities for object classification tasks such as semantic segmentation \cite{krizhevsky2012imagenet}. For CNNs, as well as for other deep learning architectures, crucial requirements for real-world applications are that the learned classifiers (a) make accurate predictions and (b) that the systems' decision making is transparent and comprehensible to humans \cite{Ribeiro2016,muggleton2018ultra}. Explanations of a system's decision making process can help machine learning experts to uncover unwanted biases. Additionally, for domain experts without a background in machine learning, explanations are crucial for being able to understand and trust the propositions of a classifier \cite{muggleton2018ultra}. Applications in the medical or pharmaceutical fields particularly require the trust of the end user, since a physician will not trust the decision of a black-box unless this decision is comprehensible

In the context of image classifications, many approaches for visual explanations have been proposed \cite{zhang2018visual}, such as LRP (Layer-wise Relevance Propagation, \cite{bach2015pixel}) or LIME (Local Interpretable Model-Agnostic Explanations, \cite{Ribeiro2016}). For explaining image classifications, LIME relies on segmentation of the image into superpixels, that is on similarity based grouping of pixels into larger structures based on local features \cite{fulkerson2009class}. LIME by default applies the specific superpixel algorithm Quick-Shift. The segmentation of an image into superpixels is crucial for the generation of the explanation in LIME since perturbation of superpixels is used to identify which of the image areas has been relevant for a specific class decision. Therefore, we were interested in exploring whether different superpixel approaches have a significant impact on the kind of visual explanations generated by LIME. Furthermore -- in the case of differences between the superpixel approaches -- it is also of interest how similar these results are to reference assessments generated by a humans based on relevance. 

As an application domain we focus on biological data which come in a huge variety of image types -- from fine grained microscopic images to holistic images of plants and animals. Our two case studies focused on applications from the medical and biological field, namely the detection of malaria parasites in thin blood smear images \cite{rajaraman2018pre} and the detection of stress in tobacco plants used for pharmaceutical purposes \cite{stocker2013machine}.

In the following, we will first recapitulate the basic concepts of LIME. Afterwards, we will introduce a variety of superpixel approaches which are well known in computer vision. Furthermore, we present the malaria domain and evaluation results -- showing differences of LIME's relevance explanation for the considered  superpixel approaches and similarity to the relevance selection
. Additionally, we shortly present and discuss the tobacco domain. We conclude with a short discussion and further work to be done.

\section{Visual explainability with LIME}
LIME \cite{Ribeiro2016} is an explanation framework for the decision of any machine learning classifier. In the original implementation it is capable of processing classifiers that either have text, images or tabular data as input. In this work we focus on the explanation of decisions for classifiers that process image data. The output of LIME therefore is a set of connected pixel patches along with a weighting for each patch. These weights indicate how strong a patch is correlated with the classifier decision.

Given a classifier $f$ and an image instance $x$, LIME outputs the weights $w$ for all pixel patches $x'$ of the image $x$. $w$ can be seen as coefficients for a linear model that acts as a surrogate for the possibly complex decision boundary of $f$. This linear model $g$ should approximate the decision boundary in the locality of $x$. To achieve this, first a pool $\mathcal{Z}$ with size $N$ (user defined constant) of perturbed versions (in the following named $z'$) of $x'$ is generated. For images, this is achieved by randomly removing patches from the image and replacing them with the mean color of the patch or with some chosen color (default is grey). Every instance of $\mathcal{Z}$ consists of the triples $\langle z'_i, f(z_i), \pi_x(z_i) \rangle$ with $f(z_i)$ being the classification result of $f$ for the perturbed version $z'_i$ in the image space and $\pi_x(z_i)$ being a proximity measure that indicates how different the perturbed version is from the original instance. This measure is used to enforce locality for the linear model $g$. 
The weights $w$ are ultimately found through K-Lasso, a procedure that is based on the regression method Lasso  \cite{Tibshirani.1996}. The input is the pool $\mathcal{Z}$ and a user defined feature limit $K$ which is the number of patches the user wants in its explanation.


\section{Superpixel methods}
Pixels, which are used to represent images in grid form, do not represent a natural representation of the depicted scene. If a single image pixel is viewed, neither its origin in the original image nor its semantic meaning can be determined. This results from the process of creating digital images. Pixels are artefacts that are created by the process of taking and creating the digital image \cite{Ren.2003}.

In comparison, the origin and semantic meaning of a superpixel can be determined. A superpixel is a local grouping or combination of pixels based on common properties, such as the color value. The advantages of superpixels can be summarized as follows \cite{Ren.2003}:
\begin{itemize}
	\item Lower complexity: Although the superpixel algorithm must be applied first to enable the name-giving groupings of pixels, this process reduces the complexity of the image due to the small number of entities. In addition, subsequent steps based on these superpixels require significantly less processing power.  
	
	\item Significant entities: Individual pixels are not very meaningful. However, pixels in a superpixel group share properties such as texture or color distribution. Through this embedding, superpixels gain an expressiveness. 
	
	\item Marginal information loss: superpixel approaches tend to oversegmentation. Thus, important areas are differentiated, but also insignificant ones. However, this apparent disadvantage basically has the positive aspect of only a minor loss of information.  
\end{itemize}
\subsection{Felzenszwalb}
The algorithm of Felzenszwalb and Huttenloch (FSZ) \cite{Felzenszwalb.2004} is to be categorized as a graph-based approach and can be described as an edge-oriented method. The approach has a complexity of $ \mathcal{O}(M \ log \ M) $.


First, the algorithm calculates a gradient between two adjacent pixels. This is
weighted according to the characteristic properties of the pixels, for example
based on the color and brightness of the individual pixels. Subsequently, individual segments - the seed for future superpixels - are formed per pixel. The aim of this process is to make the differences between the gradients within the segment as small as possible but make the differences as large as possible for adjacent segments.
The resulting superpixels should neither be too small or too large. 
However, this algorithm lacks a direct influence on the size and number
of superpixels. This usually results in a very irregular size and shape distribution \cite{Felzenszwalb.2004}.


\subsection{Quick-Shift}
Quick-Shift (QS)  is an algorithm LIME uses by default, it is described in detail in \cite{Vedaldi.2008}. Its uses a so-called \textit{mode-seeking} segmentation scheme to generate superpixels. This approach moves each point $ x_i $ to the next point which are higher density ($ P $), which causes an increase in the density. QS does not have the possibility of controlling neither the number nor the size of the superpixels. 

\subsection{SLIC}
 As the name Simple Linear Iterative Clustering (SLIC) \cite{Achanta.2012} suggests, this superpixel algorithm belongs to the group of cluster-based algorithms. SLIC uses the well-known K-Means algorithm \cite{Lloyd.1982} as a basis, but there are essential differences:
\begin{itemize}
	\item The search space ($ 2S \times 2S $) is limited proportional to the size of the superpixel ($ S \times S $). This significantly reduces the number of distance calculations. \item In addition, the complexity is independent of the number of superpixels $ k $, whereby SLIC has a complexity of $ \mathcal{O}(N) $.
	\item Furthermore, a weighted distance measure (see equation \ref{eq:slic_distance}) combines the spatial ($ d_s $) and color ($ d_c $) proximity. 
	\item In addition, the control of compactness and size of the superpixels is ensured by a parameter ($ m $).
\end{itemize}
\begin{equation}
\label{eq:slic_distance}
D = \sqrt{d_c^2 + (\dfrac{d_s}{S})^2m^2}
\end{equation}
With the parameters $ k $ the desired number of superpixels is defined. The cluster process starts with the initialization of $ k $ cluster centers (mathematically: $ C_k = [l_k,a_k,b_k,x_k,y_k]^T $), which are scanned by a regular grid with a distance of $ S $ pixels. By $ S = \sqrt{N/k} $ approximately even superpixels are guaranteed. Next, the centers are shifted in the direction of the position of the smallest gradient within a $3 \times $3 range. This is done, among other things, to avoid placing a superpixel at an edge.  

Then each pixel $ i $ is assigned to the nearest cluster center whose search area ($ 2S \times 2S $) overlaps with the position of the superpixel ($ S \times S $). The nearest cluster center is determined by the distance measure $ D $ (see equation \ref{eq:slic_distance}).

Then the average $ [l\ a\ b\ x\ y]^T $ vector of the pixels belonging to each cluster center is calculated by an update step for each cluster center and adopted as the new cluster center. Finally, a residual error $ E $ between the new and the old cluster center is determined. The assignment and calculation step can be repeated until the residual error reaches a threshold value ($ E \leq Threshold$). Finally, all unconnected pixels are added to a nearby superpixel.
\subsection{Compact-Watershed}
The Compact-Watershed (CW) \cite{Neubert.2014} algorithm is an optimized -- respectively a more compact -- version of the superpixel algorithm Watershed \cite{Meyer.1992}. 
As input a gradient image is used. Because the grey-tone of each pixel is considered as an altitude, the input can be seen as a topographical surface. Then this surface gets continuously flooded, resulting in watershed with catchment basins. During this process over-segmentation may occur. For prevention, so called markers are used \cite{Meyer.1992}.: 
\begin{enumerate}
	\item The set of markers(for each one a different label) where the flooding should begin has to be chosen.
	\item A priority queue will be created and collects the neighboring pixels of each marked area. Each pixel is graded a priority level which corresponds with the gradient magnitude of the pixel. 
	\item The pixel with the highest level of priority, gets pulled out of the priority queue. This pixel gets labeled with the same label as its neighbors if all of its neighbors are already labeled. The neighbor pixels who are not yet marked and are not contained in the priority queue are pushed into this queue. 
	\item Repeat the previous step (3) until the priority queue is empty.
\end{enumerate}
Those pixel who are still not labeled after the priority queue is empty are the watershed lines. 


Compact-Watershed is derived from the original Watershed algorithm resulting in more compact superpixels in terms of size and extension.
This is achieved by using a weighted distance measure between Euclidean distance of a pixel from the
superpixel's seed point and the difference of the pixel's grey value compared to the seed pixel's grey value.

\begin{figure}
	\centering
	\begin{subfigure}[t]{0.45\linewidth}
		\centering
		\includegraphics[width=0.6\linewidth]{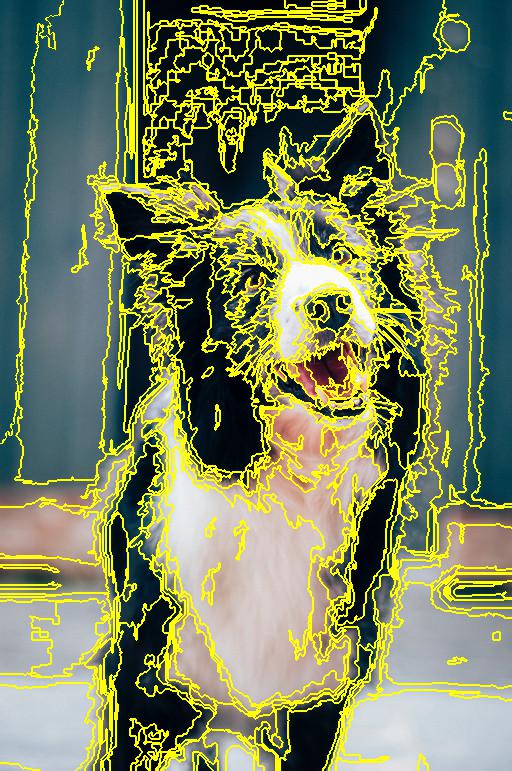}
		\caption{Felzenszwalb}
	\end{subfigure}
	\begin{subfigure}[t]{0.45\linewidth}
		\centering
		\includegraphics[width=0.6\linewidth]{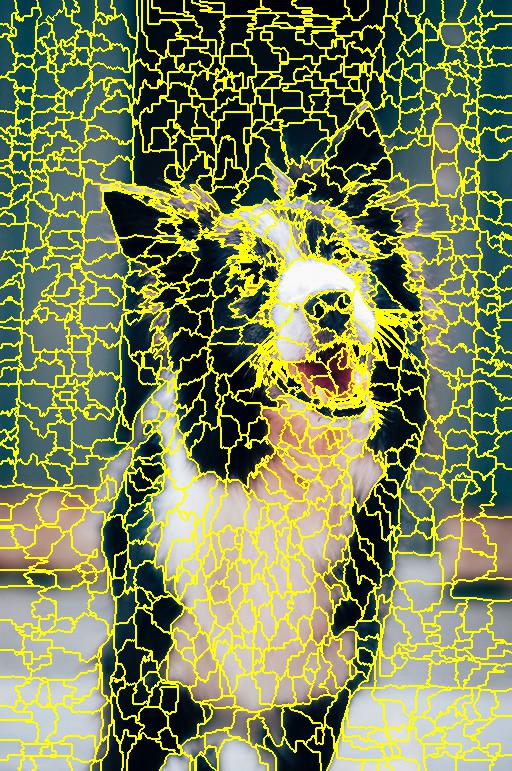}
		\caption{Quickshift}
	\end{subfigure}
	
	\begin{subfigure}[t]{0.45\linewidth}
		\centering
		\includegraphics[width=0.6\linewidth]{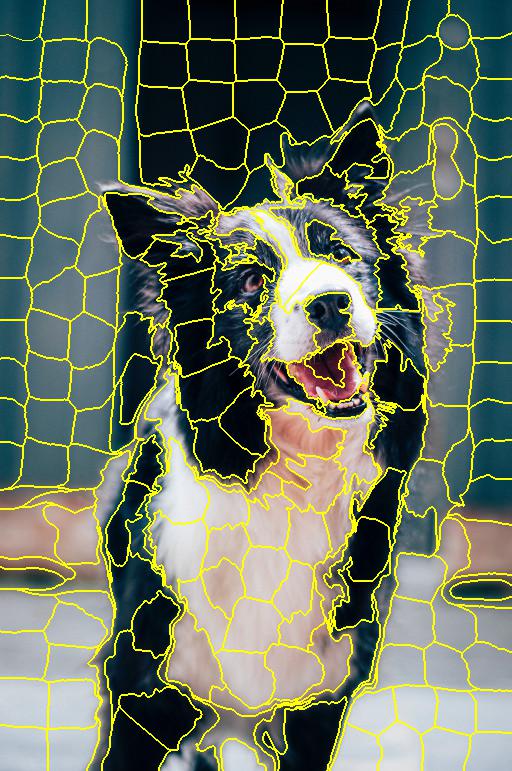}
		\caption{SLIC}
	\end{subfigure}
	\begin{subfigure}[t]{0.45\linewidth}
		\centering
		\includegraphics[width=0.6\linewidth]{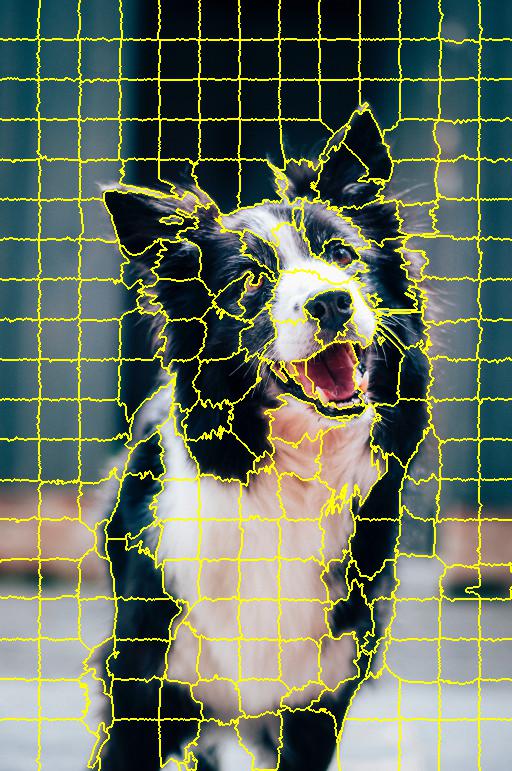}
		\caption{Compact-Watershed}
	\end{subfigure}		
	\caption{Superpixel approaches in comparison}
	\label{fig:superpixel:segmentations}
	\source{Original Photo by Baptist Standaert on Unsplash}
\end{figure}
\section{Case Studies}

\subsection{Malaria}
	Malaria is a parasitic infectious disease. It is predominantly transmitted by anopheles mosquitoes, but can also be transmitted from person to person. This happens for example by blood transfusion, organ transplantation or by sharing injection needles \cite{Nocht.1936,WorldHealthOrganization.2018}. Malaria killed 435,000 people in 2017.  Of these 266,000 were children under 5 years of age
	\cite{WorldHealthOrganization.2018}. 
	
A network trained for the detection of malaria in cells and whose results are comprehensible by LIME thus has a great benefit in the application in the field of diagnosis of malaria. For this purpose a ResNet50 \cite{KaimingHe.2015} was trained, the results are shown in Table \ref{tab:erg:malaria}.

The malaria data set \cite{rajaraman2018pre} consists of blood smear images of the most used diagnostic tool {Rapid Diagnostic Tests} (RDT) \cite{WorldHealthOrganization.2018}. The data are divided into two classes: positive and negative malaria labeled cells. In particular, the relatively large number and equally distributed (50\%-50\%)   of training examples (26,758 total) promise a good basis for a meaningful network to assess whether a cell is infected with malaria or not.
\begin{table}
	\centering
	\caption{Model results for the Malaria model }
	\label{tab:erg:malaria}
	\begin{tabular}{|l|l|}
		\hline
		Metric & Value \\ \hline
		Training accuracy  & 97.8182\%           \\
		Training loss (cross entropy)         & 0.0573            \\
		Validation accuracy & 96.5167\%            \\
		Validation loss (cross entropy)      & 0.0970            \\
		Test accuracy       & 96.3715\%           \\
		Test loss (cross entropy)             & 0.1069         \\ \hline
	\end{tabular}
\end{table}

\begin{figure}
	\centering
\begin{subfigure}[t]{0.15\textwidth}
		\centering
		\includegraphics[width=\textwidth]{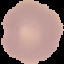}
		\caption{\textit{non-infected} cell \label{fig:malaria:good}}
	\end{subfigure}
\begin{subfigure}[t]{0.15\textwidth}
		\centering
		\includegraphics[width=\textwidth]{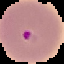}
		\caption{\textit{infected} cell \label{fig:malaria:bad}}
\end{subfigure}
	\caption{Examples of the malaria dataset}
\end{figure}

\subsubsection{Experiments} To enable an objective comparison of the superpixel approaches, the Jaccard-Coefficient \cite{Jaccard.1902} was calculated, which indicates the similarity of two sets. The similarity measure is determined between the results of the different superpixel approaches and the respective average relevant area of the decision. The relevant area per image is selected manually selects the indicator, which is most relevant to the decision making process. 

\begin{figure}
	\centering
	\begin{subfigure}[t]{0.15\textwidth}
		\centering
		\includegraphics[width=\linewidth]{./org}
		\caption{Original}	
	\end{subfigure}
	\begin{subfigure}[t]{0.15\textwidth}
		\centering
		\includegraphics[width=\linewidth]{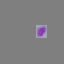}
		\caption{Average selection}	
	\end{subfigure}
	\caption[Original and Average selection of a blood smear image]{Original blood smear image from the malaria data set, as well as the corresponding average selection}
	\label{fig:erg:expertchoice}
\end{figure}

\begin{figure}
	\centering
	\begin{subfigure}[t]{0.16\textwidth}
		\centering
		\includegraphics[width=0.8\linewidth]{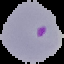}
	\end{subfigure}
	\begin{subfigure}[t]{0.16\textwidth}
		\centering
		\includegraphics[width=0.8\linewidth]{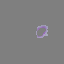}
	\end{subfigure}	
	\begin{subfigure}[t]{0.16\textwidth}
		\centering
		\includegraphics[width=0.8\linewidth]{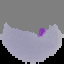}
	\end{subfigure}
	\begin{subfigure}[t]{0.16\textwidth}
		\centering
		\includegraphics[width=0.8\linewidth]{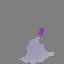}
	\end{subfigure}		
	\begin{subfigure}[t]{0.16\textwidth}
		\centering
		\includegraphics[width=0.8\textwidth]{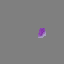}
	\end{subfigure}
	\begin{subfigure}[t]{0.16\textwidth}
		\centering
		\includegraphics[width=0.8\textwidth]{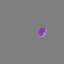}
	\end{subfigure}
	
	\begin{subfigure}[t]{0.16\textwidth}
		\centering
		\includegraphics[width=0.8\linewidth]{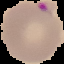}	
		\caption{Original}	
	\end{subfigure}
	\begin{subfigure}[t]{0.16\textwidth}
		\centering
		\includegraphics[width=0.8\linewidth]{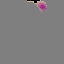}	
		\caption{Felzen-\\szwalb}
	\end{subfigure}
	\begin{subfigure}[t]{0.16\textwidth}
		\centering
		\includegraphics[width=0.8\linewidth]{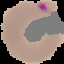}
		\caption{Quick-\\Shift}
	\end{subfigure}	
	\begin{subfigure}[t]{0.16\textwidth}
		\centering
		\includegraphics[width=0.8\linewidth]{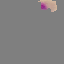}
		\caption{Quick-\\Shift opt.}
	\end{subfigure}
	\begin{subfigure}[t]{0.16\textwidth}
		\centering
		\includegraphics[width=0.8\textwidth]{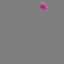}
		\caption{SLIC}	
	\end{subfigure}
	\begin{subfigure}[t]{0.16\textwidth}
		\centering
		\includegraphics[width=0.8\textwidth]{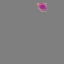}
		\caption{Compact-Watershed}
	\end{subfigure}
	\caption{LIME results for true positive predicted malaria infected cells}
	\label{fig:erg:lime:malariatrue}
\end{figure}

For the comparability of the results, 100 images of infected cells were selected from the test data set of the malaria blood smear images. To make it easier to select, only images with a single malaria indicator were selected. Of these 100 images, 85 were classified by the network as \textit{infected} (\textit{true positive}). The remaining 15 were classified as \textit{not-infected} (false negative).  Table \ref{tab:erg:mean} shows the result of the Jaccard-Coefficient for the respective superpixel approach with the \textit{true positive} classified explanations, where only the most important feature for the network's decision (see Figure \ref{fig:erg:lime:malariatrue}) is displayed. All of the superpixel methods were optimized to the given case to maximize the average Jaccard-Coefficient
, hence the optimized Quick-Shift version. This was done so that all of the superpixel approaches would be compared on a fair level.  

\begin{table} 
\caption{Jaccard Coeffficient of the different superpixel methods}\label{tab:erg:mean}
\centering
\begin{tabular}{|l|l|l|l|}
\hline
Superpixel method   & Mean Value   &  Variance & Standard deviation  \\ 
\hline
Felzenszwalb         & 0.85603243 & 0.03330687 & 0.18250170 \\
Quick-Shift           & 0.52272303 & 0.04613085 & 0.21478094 \\
Quick-Shift optimized & 0.88820585 & 0.00307818 & 0.05548137 \\
SLIC                 & 0.96437629 & 0.00014387 & 0.01199452 \\
Compact-Watershed    & \textbf{0.97850773} & \textbf{0.00003847} & \textbf{0.00620228} \\ 
\hline
\end{tabular}
\end{table}

\begin{figure}
	\begin{subfigure}[t]{0.16\textwidth}
		\centering
		\includegraphics[width=0.8\linewidth]{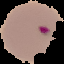}
		\caption{Original}
	\end{subfigure}
	\begin{subfigure}[t]{0.16\textwidth}
		\centering
		\includegraphics[width=0.8\linewidth]{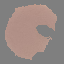}
		\caption{Felzen-\\szwalb}
	\end{subfigure}
	\begin{subfigure}[t]{0.16\textwidth}
		\centering
		\includegraphics[width=0.8\linewidth]{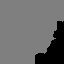}
		\caption{Quick-\\Shift}
	\end{subfigure}	
	\begin{subfigure}[t]{0.16\textwidth}
		\centering
		\includegraphics[width=0.8\linewidth]{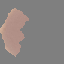}
		\caption{Quick-\\Shift opt.}				
	\end{subfigure}	
	\begin{subfigure}[t]{0.16\linewidth}
		\centering
		\includegraphics[width=0.8\textwidth]{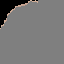}	
		\caption{SLIC}
	\end{subfigure}
	\begin{subfigure}[t]{0.16\linewidth}
		\centering
		\includegraphics[width=0.8\textwidth]{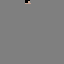}
		\caption{Compact-Watershed}
	\end{subfigure}
	\caption{LIME results for false positive predicted malaria infected cells}
	\label{fig:erg:lime:malariabad}
\end{figure}


\subsection{Tobacco plants}
Tobacco is a significant plant used in biopharmaceutical production using genetically modified (GM) plants. Two important reasons are its ability to produce biomass quickly and minimum risk of food chain contamination because of the fact that tobacco is not a food crop \cite{tremblay2010tobacco}. It is able to produce proteins which can be used for treatment or diagnosis of various diseases. However, if plants are used to produce medicine for human use, strict regulations present in the field of pharmaceutical production must be observed. In this context it is desirable to monitor the health state of each plant to ensure only healthy plants are used for drug production, however different parts of world regulate pharmaceutical production of GM plants differently \cite{spok2008evolution}.
\begin{figure}
	\centering
	\begin{subfigure}[t]{0.25\textwidth}
		\centering
		\includegraphics[width=\textwidth]{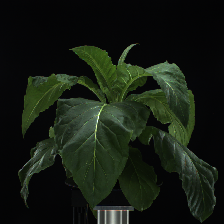}
		\caption{Healthy Tobacco plant\label{fig:tabak:gesund}}
	\end{subfigure} 
	\begin{subfigure}[t]{0.25\textwidth}
		\centering
		\includegraphics[width=\textwidth]{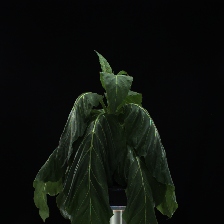}
		\caption{Stressed Tobacco plant\label{fig:tabak:gestresst}}
	\end{subfigure}
	\caption{Examples of the Tobacco plant dataset}
	\label{fig:tabak}
\end{figure}

Stocker, et al. have investigated various methods to classify stress in tobacco plants using non neuronal AI approaches \cite{stocker2013machine}. Figure \ref{fig:tabak} shows sample images of healthy and stressed tobacco plants. We use the same tobacco data set as a case study to assess the suitability of CNNs for stress classification again using LIME to provide insights into the classification process. For the training the same ResNet50 as for the malaria dataset was used. The only difference was that, to prevent overfitting, the last layers were unfrozen during the training of the tobacco trainingsset.  
The tobacco data set consists of 700 images total divided into two classes, healthy and stressed. Only 81 images of stressed plants are contained in the data set, so expectations of a good classification result were limited.

Table \ref{table:erg:model:Tabakpflanzen} shows the trained model results on the tobacco plant data set. These clearly already show that the results should not be trusted to begin with, so we decided to discontinue work on this case study for the time being. 
\begin{table} 
	\centering
	\caption{Model results for Tobacco plants}
	\label{table:erg:model:Tabakpflanzen}
	\begin{tabular}{|l|l|}
		\hline
		Metric & Value \\ \hline
		Training accuracy  & 91.2577\%           \\
		Training loss         & 0.4459             \\
		Test accuracy       & 50\%           \\
		Test loss              & 0.7524          \\ \hline
	\end{tabular}
\end{table}

\begin{table}[!htb]
	\centering
	\caption{Hyperparameter for the training of both models}
	\label{table:erg:model:hyperparameter}
	\begin{tabular}{|l|l|}
		\hline
		Hyperparameter & Value \\  \hline
		Epochs & 50 \\
		Batch size & 32 \\  
		SGD learningrate & 0,0001 \\ 
		SGD Momentum & 0,90 \\ 
		SGD Nestrov & Ja \\ 
		Dropout & 0,50 \\ 
		L2-Regulation & 0,0001 \\		 \hline
	\end{tabular} 
\end{table}
\section{Discussion}
In the following, the experiments with LIME from the previous chapter, their results and the possible improvement of the visual explainability are discussed (see Table \ref{tab:erg:mean}). 
Although {FSZ} is an older algorithm compared to the other approaches considered, still good results were achieved. Surprisingly, \textit{QS}, the standard algorithm of LIME, is surpassed by about 33.33\%. Since FSZ itself does not have any parameter, which could limit the size of a superpixel, it seems that LIME can act pretty much freely and can generate the superpixels purely based on relevance regarding explainability. A good example for such a case is the explanation from LIME for the false positive classification shown in Figure \ref{fig:erg:lime:malariabad}. In contrast to the other superpixel methods explored though FSZ's decision for \textit{not infected} is more comprehensible. However the variance and  standard deviation for the true positive examples, indicates the similarity vary significantly and with FSZ the results are not stable and may  sometimes show regions as relevant for the decision which are actually not important. This for example is the case for the first result from LIME while using FSZ (see Figure \ref{fig:erg:lime:malariatrue}). 

The optimized version of {QS}, remarkably achieved an improvement of 36.55\% compared to the standard version of LIME. Additionaly it performs slightly better than {FSZ} - with an improvement of 3.22\% - and the variance and standard deviation are also lower, which indicates the results are more stable than with {FSZ} and the unoptimized {QS} version. 

{SLIC} makes it possible to influence the actual size of the superpixels through a parameter. Consequently, the higher similarity measure with over  44.17\% compared to {QS} and over  7.62\% compared to the optimized version of {QS}, is not surprising. Additionally, a lower variance and standard deviation was achieved. These results show that {SLIC} has advantages over {QS} due to showing a better correspondence between superpixels and relevant areas. 

The last superpixel approach compared with {QS} was {CW}. Like {SLIC} it supports  influencing the compactness of the resulting superpixels. In comparison to all other superpixel approaches {CW} yielded the best results. This approach achieved an improvement of  45.58\% over the standard {QS} and  an improvement of 9.03\% over the optimized {QS} version. It also significantly reduces the variance
 and standard deviation
. This indicates there is a very correspondence over all the 85 images.

\section{Conclusion}
Our results suggest that tailoring of the superpixel
approach - whether by an optimized version of QS or by FSZ,
SLIC or CW - to the task will improve the visual explainability of LIME. Therefore a selecting a suitable algorithm for LIME can be beneficial and should be considered. With the exception of {QS} the remaining approaches segment fewer irrelevant areas of an image (see Figure \ref{fig:erg:lime:malariatrue}). It was also observed that {CW} achieved the best results.

In applications where large area and uneven features are to be emphasized, an approach like {CW} would possibly do worse because it divides the input into very small, evenly sized superpixels.  {FSZ}, which generates superpixels in significantly different sizes, may even achieve the best results in such application areas. Consequently, the finding that {CW} does give the best results in malaria is not universally valid and the superpixel approaches should be evaluated by experts in different application areas. Another conclusion is that superpixel methods other than {QS} are more suitable for LIME.  

Since the area of pharmaceutical and agricultural applications is an emerging research area for applying machine learning to digital  plant phenotyping tasks, we plan to continue pursuing the ideas begun in the tobacco case study. We suspect that an objective assessment of plant health will yield better results if based on 3D data, because the habitus of a plant should then be represented more realistically than in a purely texture based 2D analysis as in the tobacco case study. Furthermore, the number of training images in said case study was insufficient, so the goal will be to generate a greater data set containing 3D scans of plants to continue research on this subject.

%
%
%
%
\bibliographystyle{splncs04}
\bibliography{bibliographie}
%
%
%
%
\end{document}